\documentclass[11pt,letterpaper]{article}
\usepackage[pass]{geometry}
\usepackage[hyperref]{acl2017}
\usepackage{times}
\usepackage{latexsym}
\usepackage[utf8]{inputenc}
\usepackage{amsmath}
\usepackage{amssymb}
\usepackage{url}
\usepackage{booktabs}
\usepackage{graphicx}

\aclfinalcopy 

\title{Overview of the Ugglan Entity Discovery and Linking System}

\author{Marcus Klang \\
\\
 \\
 \\
  {\tt marcus.klang@cs.lth.se} \\\And
  Firas Dib \\
  Lund University\\
  Department of Computer Science \\
  S-221 00 Lund, Sweden \\ \\
    {\tt firas.dib@gmail.com} \\\And
  Pierre Nugues \\
\\
 \\
 \\
  {\tt pierre.nugues@cs.lth.se} \\}

\date{}

\begin{document}
\begin{small}
\maketitle
\end{small}
\begin{abstract}
\end{abstract}
Ugglan is a system designed to discover named entities and link them to unique identifiers in a knowledge base.
It is based on a combination of a name and nominal dictionary derived from Wikipedia and Wikidata, a named entity recognition module (NER) using fixed ordinally-forgetting encoding (FOFE) trained on the TAC EDL data from 2014-2016,  a candidate generation module from the Wikipedia link graph across multiple editions, a PageRank link and cooccurrence graph disambiguator, and finally a reranker trained on the TAC EDL 2015-2016 data.

In our first participation in the TAC trilingual entity discovery and linking task, we obtained a \textit{strong typed mention match} of 0.701 (Ugglan2), a \textit{strong typed all match} of 0.592 (Ugglan4), and \textit{typed mention ceaf} of 0.595 (Ugglan1).

\section{Introduction}
The goal of the trilingual entity discovery and linking task (EDL) in TAC 2017 \citep{Ji2017} is to recognize mentions of entities in Chinese, English, and Spanish text and link them to unique identifiers in the Freebase knowledge base. In the TAC datasets, the mentions have a type, either persons (PER), geopolitical entities (GPE), organizations (ORG), locations (LOC), or facilities (FAC), and their syntactic form can consist of proper or common nouns, called respectively named and nominal mentions. Some entities in the annotated corpus are not in Freebase. They are then linked to a NIL tag and clustered across the three languages; each specific entity being assigned a unique identifier.


In this paper, we describe Ugglan, a generic multilingual EDL platform that required minimal adaptation to the TAC 2017 tasks. We detail the system architecture and its components as well as the experimental results we obtained with it.

\section{System Overview}
Ugglan has a pipeline architecture that consists of three main parts:
\begin{itemize}
\item A mention discovery that uses a finite-state automaton derived from Wikipedia and/or a feed-forward neural network trained on the TAC 2014-2016 data \citep{Ji2014,Ji2015,Ji2016b};
\item An entity linker that uses mention-entity pairs extracted from Wikipedia and ranks them using PageRank;
\item A reranker trained on the TAC 2014-2016 data.
\end{itemize}

The Ugglan architecture is modular and parameterizable, and its parts can use independent algorithms. To build it, we used a set of resources consisting of Wikipedia, Wikidata, and DBpedia.

\subsection{Mention Discovery}
The first part of the processing pipeline is the discovery of mentions of entities in text. It starts with a custom multilingual rule-based tokenization of the text and a sentence segmentation. We then normalize the letter case based on statistics from all the Wikipedia pages.

We discover the mentions using a combination of a finite-state transducer (FST) built from mentions extracted from Wikipedia and an optional named entity recognizer (NER) based on neural networks. As an alternative, Ugglan can also use the Stanford NER \citep{Finkel2005}.

\begin{figure}[th]
\begin{center}
\includegraphics[width=0.7\columnwidth]{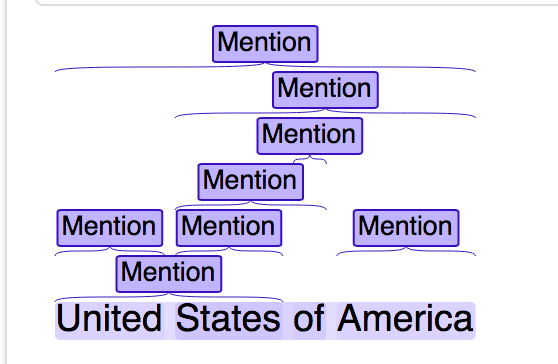}
\end{center}
\caption{Mention candidates produced by the finite-state transducer for the phrase \textit{United States of America}}
\label{fig:us_mentions}
\end{figure}

The mention discovery results in an overgeneration of mention candidates. For instance, the phrase 
\begin{quote}
United States of America
\end{quote}
results into as many as eight candidates; see Fig.~\ref{fig:us_mentions}. We prune them using parameterized rules. 

\begin{figure*}
  \centering
  \includegraphics[width=\textwidth]{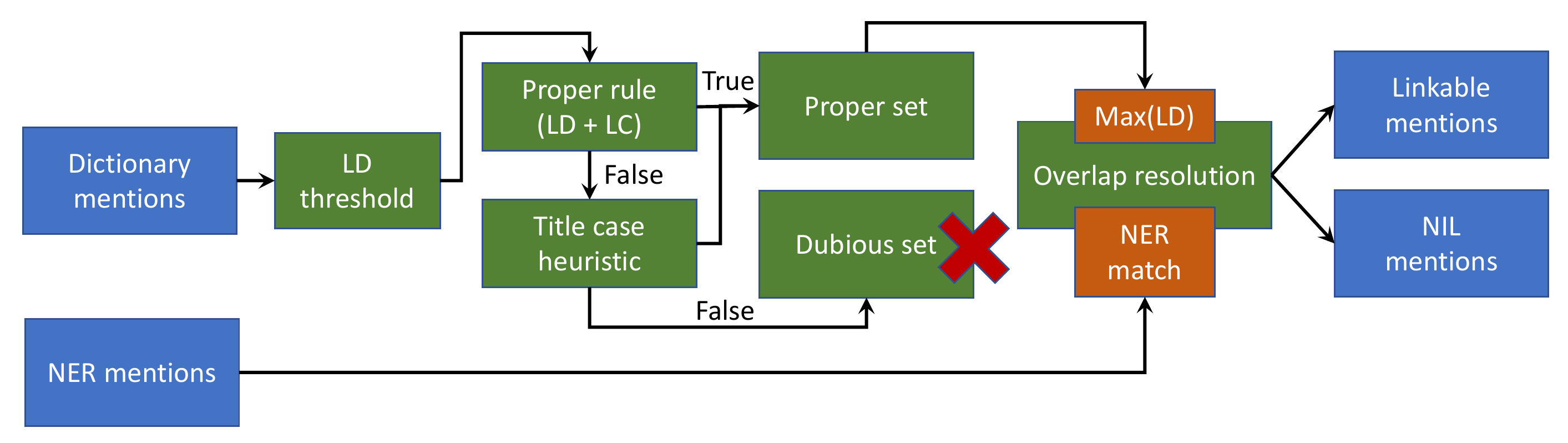}
  \caption{Overview of the mention pipeline}
  \label{figure:mention-pipeline}
\end{figure*}

Finally, our system does not output overlapping mentions. We resolve this overlap using a statistical estimation of the mention ``linkability'': The \textbf{link density}; see Sect.~\ref{sec:filtering}. Figure~\ref{figure:mention-pipeline} shows the overall mention detection steps.

\subsection{Entity Linking}
Once we have carried out the detection, we associate each mention with entity candidates by querying a mention-entity graph. 

Some of the entities are referred by mention variants along a text, for instance starting with ``Barack Obama'' and then ``Obama'' or ``Barack''. We augment the entity recall by sorting the mentions in a document with a partial ordering using the \verb=is_prefix= or \verb=is_suffix= relations so that we have:
\[
\begin{array}{lcll}
\text{Barack} &\prec& \text{Barack Obama}& \text{and}\\
\text{Obama} &\prec &\text{Barack Obama}.
\end{array}
\]
Using this ordering, we expand the candidates of the preceding strings by adding the candidates of the including strings further up in the order. For instance, we add all the candidate entities of ``Barack Obama'' to the candidates of ``Obama''.

We extracted the graph of mentions to candidate entities from Wikipedia as well as the graph of entity--entity cooccurrences. We built this graph from outlinks gathered from the combination of seven Wikipedia editions.

Finally, we disambiguate the text entities using a local graph of candidates, on which we apply the PageRank algorithm. For each mention, PageRank assigns a weight to the candidates that enables us to rank the entities.

\subsection{Reranking and Classification}
After the entity linking step, each mention has a ranked list of entity candidates. We rerank these lists using a multilayer neural network trained on the TAC2015-16 data. This also results in some entities being assigned the NIL identifier.

We assign a type to the entities using a predefined dictionary mapping derived from DBpedia; this type is possibly merged with that obtained from the NER, if no available mapping exists.

At this point, we have discovered and resolved the named expressions. We apply a discovery to the nominal expressions (NOM) using a dictionary collected from Wikidata and a coreference resolution based on exact string matches.

Finally, we discard the classes not relevant to the TAC task.

\section{Building Ugglan's Knowledge Base}
Ugglan relies on a graph of mention-entity and entity-entity for both the discovery and the linking stages. We constructed this knowledge base from a set of resources:
\begin{itemize}
    \item Seven Wikipedia editions: en, es, zh, de, fr, ru, and sv;
    \item Wikidata, which binds these editions together;
    \item DBPedia, which we used for the class mapping.
\end{itemize}

\subsection{The Wikipedia Corpus}
The Wikipedia corpus is our first resource from which we extract the text, the mentions, and link graph. We can access its content in multiple ways and with varying degrees of accuracy. 

A simple approach is to download a XML dump, apply some filtering to the raw wiki markup, and then use this output as text. However, Wikipedia features many templates that are language dependent. Using a dump approach would leave the template expansion unsolved, as well as the local context of the links and information about the logical structure. 

We opted therefore to use a rendered HTML version which expands the templates and has its Scribuntu scripts executed. This produces a HTML dump of Wikipedia, which is as true to the user online version as possible without actually replicating the full Wikipedia infrastructure. Concretely, we used a combination of Xowa (offline Wikipedia reader) and the public Wikipedia REST API to get the HTML dumps.

\subsubsection{HTML Annotation Processing}
The raw HTML must be filtered and simplified to be easy to process. We converted the hierarchical structure using a DOM tree produced by Jsoup and applied heuristics to produce a flattened version with multiple layers of annotations using the Docforia structure \citep{klang2017a}. These layers include information on anchors, headers, paragraphs, sections, lists, tables, etc.

The Chinese version was processed in a special way as it can be translated in multiple variants: simplified, traditional, and localizations. To get coherent statistics, we reimplemented the translation mechanism used by the online version to produce a materialized \verb=zh-cn= Wikipedia edition.

We finally resolved all the pages and anchors in the flattened version to Wikidata, which produces multilingual connections for most entities.

\subsection{Multilingual Resources: Wikidata and DBpedia}
Wikidata is Ugglan's repository of multilingual entities as it contains clean mappings between the multiple language editions, as well as detailed structured data. TAC uses Freebase as knowledge base and we created mappings to translate Freebase entities to Wikidata. Wikidata entities are represented by unique identifiers called Q-numbers. We mapped the Wikidata entities missing from Freebase to NIL-xx identifiers, where xx is a number unique for the entity.

We chose DBpedia to map entities to TAC classes as it produced subjectively better mappings than using Wikidata. Wikidata would have required additional rules to carry out the conversion.

\subsection{Mention--Entity Graph}
\label{sec:mention-entity}
We associated each mention in Wikipedia with a set of potential entities in text. We used a dictionary, where the entry (or key) is the entity mention from Wikipedia and the value is the Q-number. In Wikipedia, we extracted the mentions from five sources across the languages we consider:

\begin{description}
\item[The title] of the entity's Wikipedia page with and without parentheses. For instance, we have $M_{Q90} = \{\text{Paris}\}$ and $M_{Q167646}$ = \{\text{Paris, Paris\_(mythology)}\}.


\item[The Wikipedia redirects:] i.e. page names that automatically redirect to another page like the \textit{EU} page that moves the reader to \textit{European Union}. This enables us to collect alternative names or spelling variants so that we expand the mention list for the European Union entity, $M_{Q458}$, from  $\{\text{European Union}\}$ to $\{\text{European Union}, \text{EU}\}$.

\item[The disambiguation pages,] where a page title is associated to two or more entities. The English Wikipedia has a disambiguation page associated with \textit{Paris} that links to about 100 entities, ranging from the capital of France to towns in Canada and Denmark as well as song and film titles; 

\item[The wikilinks.] A link in Wikipedia text is made of a word or a phrase, called the label, that shows in the text, and the page (entity) it will link to, where the wiki markup syntax uses double brackets:  \verb=[[link|label]]=.  When the link and the label are different, the label is often a paraphrase of the term. We therefore consider all these labels as candidate mentions of the entity. Examples in Swedish of such labels for the former Swedish Prime Minister Göran Persson, Q53747, include the name itself, \textit{Göran Persson}, 468 times, \textit{Persson}, 14 times, and  \textit{Han Som Bestämmer}, `He Who Decides', one occurrence.

\item[The first bold-faced string.] We finally used the first bold-faced string in the first paragraph of an article as it often corresponds to a synonym of the title (or the title itself).
\end{description}

\subsection{Statistics}
\label{sec:stats}
During the mention gathering, we also derive statistics for a given language. Before we compute these statistics, we apply a procedure that we called \emph{autolinking}. In an article, the Wikipedia guidelines advise to link only one instance of an entity mention\footnote{\url{https://en.wikipedia.org/wiki/Wikipedia:Manual_of_Style/Linking\#Overlinking_and_underlinking}}: Normally the first one in the text. With the autolinking procedure, we link all the remaining mentions provided that we have sequences of exactly matching tokens. The statistics we collect are:

\begin{itemize}
\item The frequency of the mention string over the whole Wikipedia collection (restricted to one language);
\item The frequency of the pair (mention, entity) that we derive from the links without autolinking (only manually linked mentions);
\item The count of (entity1, entity2) pairs in a window corresponding to a paragraph and limited to 20 linked mentions. This is carried out after autolinking;
\item Capitalization statistics for all the tokens: We extract token counts for all tokens with a frequency greater than 5 and we break them down by case properties:  uppercased, lowercased, titlecased, and camelcase;
\end{itemize}

\section{Mention Recognition}
Ugglan uses a multilingual rule-based tokenizer and segmenter that we implemented using JFlex. For logographic languages such as Chinese, the tokens are equivalent to characters. The parser was customized to accept a mixture of both logographic and alphabetic text.

The tokenizer was used in conjunction with the Lucene analyzers. Lucene provides an  infrastructure consisting of common filters and normalizers for many languages, from which we use case folding, accent stripping, Unicode form normalization, and stemming. These pipelines are configurable and easy to adapt for new languages.

The mention discovery is carried out by two modules: A dictionary-based finite-state transducer and a named entity recognizer (NER) using neural networks that we trained  on the TAC data; see Sect.~\ref{sec:NER}. These two modules can work in tandem or independently. 

The mention recognition pipeline has three primary steps: discovery, filtering, and overlap resolution. In addition, the discovery pipeline can be configured to use one of three modes: NER-only, dictionary-only, and a hybrid mode. The primary difference between these modes is how the filtering and overlap resolution operates.
Figure~\ref{figure:mention-pipeline} shows an overview of these steps.

\subsection{Discovery}
Before querying the FST dictionary, we normalize the tokens in uppercase or lowercase characters for English and Spanish using statistics derived from Wikipedia. For instance, we convert \textit{BEIJING} into the title case variant \textit{Beijing}. However, due to ambiguity, we did not apply case normalization to title-cased words.

\subsection{Filtering}
\label{sec:filtering}
The mentions of named entities are likely to be linked in Wikipedia. Examining the articles, we observed that, given word sequence, the relative frequency of linkage often reveals its ambiguity level. For instance, while the word \textit{It} can refer to a novel by Stephen King, it is rarely an entity and, at the same time, rarely linked.

The \textit{link density} (LD) is a measure derived from the analysis of text linkage in Wikipedia. It loosely corresponds, as the original text is not fully linked, to the probability of a sequence of tokens being linked in the source edition.

We estimated the linkage probability by applying the FST dictionary to Wikipedia in an offline step. We counted the exact matches with the known ground truth: The anchors created by the Wikipedia editors. In addition, before counting, we added the \emph{autolinked} anchors that matched existing ones perfectly:


\begin{equation}
    \begin{aligned}
        \text{Link density}=\frac{\#\text{Anchor}}{\#\text{Text}+\#\text{Anchor}}
    \end{aligned}
\end{equation}

In addition to link density, we used the gold standard mention counts, the \textit{link count} (LC), as a measure of significance.

Before the overlap resolution, all the mentions from the dictionary are classified into either a proper set or a dubious set. The proper set consists of all the mentions which exceed LD and LC thresholds; The proper set also includes the mentions which do not exceed these thresholds if at least 75\% of the tokens in the mention are title cased.

\subsection{Overlap Resolution}
The mentions placed in the \emph{proper set} and the mentions found by the NER will be merged and resolved in the overlap resolution step. The NER module itself only outputs nonoverlapping mentions. This stage works differently depending on which mode is used:

\begin{itemize}
    \item \emph{NER-only} accepts only NER mentions and produces linkable mentions, if an exact dictionary match is found.
    \item \emph{Dictionary-only} ignores the NER mentions and solves overlapping mentions by picking the mention which has the largest LD value until no overlap exists.
    \item \emph{Hybrid} merges NER mentions with dictionary matches by trusting the NER output where applicable i.e. when multiple candidates exist, it chooses the NER output, otherwise the dictionary output.
\end{itemize}

\section{Named Entity Recognizer}
\label{sec:NER}
We developed a named entity recognizer (NER) based on a feed-forward neural network architecture and a \emph{fixed ordinally-forgetting encoding} (FOFE) \citep{Xu2017,Zhang2015}. This NER is part of the mention recognition module; see Figure~\ref{figure:mention-pipeline} for the dataflow.

The NER operates over sentences of tokens and outputs the highest probability class using a moving focus window with varying width. The focus window represents potential named entity candidates and ranges from one to seven words. A more detailed explanation of this, and why there is an upper limit, is explained in Sect.~\ref{ner:candidates}

The NER can recognize both named and nominal expressions and predict their class. The named or nominal types are just extensions to the classes.
If there were $N$ classes originally, there would be $2N$ outputs if all nominal classes were included. In the TAC2017 Ugglan system, the possible classes are:
\begin{itemize}
    \item \emph{\{PER, GPE, ORG, LOC, FAC\}-NAM} and
    \item \emph{\{PER, GPE, ORG, LOC, FAC\}-NOM}.
\end{itemize}

The NER is identical in its construction for English and Spanish, without any language specific feature engineering. However, we modified this module for Chinese since the word segmentation was not found reliable. In Chinese, we used the individual characters (logograms) as word features and none of the corresponding Latin character features. In any case, the Chinese character features would be a subset of the word features.

\subsection{The Fixed Ordinally-Forgetting Encoding}
We applied a \emph{fixed ordinally-forgetting encoding} (FOFE) \citep{Xu2017,Zhang2015} as a method of encoding variable-length contexts to fixed-length features. This encoding method can be used to model language in a suitable manner for feed-forward neural networks without compromising on context length.

The FOFE model can be seen as a weighted bag-of-words (BoW). Following the notation of \citet{Xu2017}, given a vocabulary $V$, where each word is encoded with a one-hot encoded vector and $S = w_1, w_2, w_3, ..., w_n$,  an arbitrary sequence of words, where $e_n$ is the one-hot encoded vector of the nth word in $S$, the encoding of each partial sequence $z_n$ is defined as:

\begin{equation}
\begin{aligned}
z_n = \begin{cases}
0, & \text{if n = 0}\\
\alpha \cdot z_{n-1} + e_{n}, & \text{otherwise},
\end{cases}
\end{aligned}
\end{equation}
where the  $\alpha$ constant is a weight/forgetting factor which is picked such as  $0 \leq \alpha < 1$. The result of the encoding is a vector of dimension $|V|$, whatever the size of the segment.






\citet{Zhang2015} showed that we can always recover the word sequences $T$ from their FOFE representations if $0 < \alpha \leq 0.5$ and that FOFE is almost unique for $0.5 < \alpha < 1$. \citet{Zhang2015} make the assumption that the representation is (almost) always unique in  real texts.





\subsection{Features}
The neural network uses both word and character-level features. The word features extend over parts of the sentence while character features are only applied to the focus words: The candidates for a potential entity.

\subsubsection{Word-level Features}
The word-level features use bags of words to represent the focus words and FOFE to model the focus words as well as their left and right contexts. As context, we used all the surrounding words up to a maximum distance, defined by the floating point precision limits using the FOFE $\alpha$ value as a guide.

Each word feature is used twice, both in raw text and normalized lower-case text. The FOFE features are used twice, both with and without the focus words. For the FOFE-encoded features, we used $\alpha = 0.5$.

The beginning and end of sentence are explicitly modeled with \texttt{BOS} and \texttt{EOS} tokens, which have been added to the vocabulary list.

The complete list of features is then the following:

\begin{itemize}
    \item Bag of words of the focus words;
    \item FOFE of the sentence:
    \begin{itemize}
    \item  starting from the left, excluding the focus words.
    \item starting from the left, including the focus words.
    \item starting from the right, excluding the focus words.
    \item starting from the right, including the focus words.
    \end{itemize}
\end{itemize}

This means that, in total, the system input consists of 10 different feature vectors, where five are generated from the raw text, and five generated from the lowercase text.

\subsubsection{Character-level Features}
The character-level features only model the focus words from left to right and right to left. We used two different types of character features: One that models each character and one that only models the first character of each word. We applied the FOFE encoding again as it enabled us to weight the characters and model their order. For these features, we used $\alpha = 0.8$.

In order to ensure the characters fall into an appropriate range, we encoded them with a simple modulo hash. Each characters ASCII value is normalized to be within the range 0 and 128. This limitation is reasonable since most characters of English and Spanish are in the ASCII table. The Spanish characters in the range 128:256 are confused with unaccented ASCII characters, for instance \textit{ñ} with \textit{q}.

\subsubsection{Projection Layers}
\paragraph{Characters.}
The character features are generated first as sparse one-hot encoded vectors of dimension 128 and then projected to a dense representation of lower dimension: 64. 
To project the character features, we used a randomly initialized weight matrix, which is trained as part of the network. 
This procedure produced better results than the direct input of one-hot vectors.

\paragraph{Words.} We projected the word-level features to a 256-dimension dense representation. We initialized the projection layer with two different word2vec \citep{Mikolov2013c} models that we pretrained on the en, es, and zh wikipedias. One model was trained on normalized text, while the other was trained on the raw untouched text. These are incorporated into the rest of the network and consequently trained as a part of it.

When creating the word projection layers from the word2vec models, we used the weights of the top 100,000 words. We disregarded all the other words and instead mapped them to a \emph{unknown word vector}. More specifically, for the case-sensitive projection layer, we return a \texttt{UNK} vector when we encounter a word with no embedding; if this word is equal to its normalized lower cased variant, we return a special \texttt{unk} vector instead.

\subsection{Named Entity Candidates}\label{ner:candidates}
The potential named entity candidates are produced by looping over each word in the sentence with a moving window that expands up to seven words. This exhaustively generates all the possible candidates in the sentence, which in turn produces a lot of noise. In the training process, we sample this noise to build a set of negative examples and instruct the network how to discriminate mention boundaries and invalid mentions.

The upper bound of seven was found by going through the annotations of the TAC 2016 data and seeing if there was any clear cut-off where results would start to diminish. After seven words, we found there was little benefit to go any further. This upper limit value is significant because each candidate which is not a positive sample is considered negative and in turn used in the training process.

A large focus window results in many negative samples, which are not representative of the real world. As the negative candidates are randomly sampled, we would (to some degree) get a skewed distribution of the negative samples. If, for example, the upper bound was set to 12 words, there would be many negative 12-token long samples in comparison to how many positive examples there are. We attempted to weigh the different selections with respect to the positive mention count. We set it aside for the TAC 2017 evaluation due to a lack of time.

In total, we considered three different cases to create the training data:

\begin{enumerate}
    \item The mention candidate matches exactly an annotated sequence;
    \item The candidate is completely disjoint, i.e., contains no annotated words;
    \item The candidate partially or completely overlaps with an annotated sequence or is a subset of the sequence,
\end{enumerate}
where an annotated sequence corresponds to all the words annotated with a given class in the data, such as \texttt{University of Lund}.

The first case corresponds to the positive examples that we label with the TAC classes, while the two last cases are the negative examples that we label as \texttt{NONE}. We will keep this stratification in the training step.


\subsection{Training}
In the data set we collected, the negative samples outnumber the positive ones by an order of magnitude. We used a manually-specified distribution of the samples to mitigate this unbalance and train the network. At the beginning of each epoch, the data is shuffled and the negative samples are re-selected according to the distribution. This means that we continuously introduce new negative examples and previously unseen data to the network, which helps with regularization.

Table~\ref{table:erd_distribution} shows the distribution we used for the TAC 2016 EDL task.

\begin{table}[ht]
  \centering
  \begin{tabular}{ l  c   }
    \hline
    \textbf{Candidate type} & \textbf{Ratio of sample size} \\
    \hline
    Negative: Overlapping & 60\% \\
    Negative: Disjoint & 30\% \\
    Positive & 10\%\\
    \hline
  \end{tabular}
  \caption{The distribution between positive and negative mention candidates.}
  \label{table:erd_distribution}
\end{table}

We trained the NER system with data from TAC 2014-15 and evaluated it on the 2016 data; see Table~\ref{table:2016_result}.

\begin{table*}[t]
  \centering
  \begin{tabular}{l | lll | lll | lll}
  \toprule 
    \textbf{Language} & \multicolumn{3}{c}{Named} & \multicolumn{3}{c}{Nominal} & \multicolumn{3}{c}{Overall} \\

    & P & R & F1
    & P & R & F1
    & P & R & F1  \\
    \midrule
    English & 0.734 & 0.816 & 0.773 & 0.580 & 0.805 & 0.674 & 0.801 & 0.676 & 0.733 \\
    Chinese & 0.769 & 0.792 & 0.780 & 0.554 & 0.757 & 0.639 & 0.769 & 0.612 & 0.682 \\
    Spanish & 0.736 & 0.685 & 0.709 & 0.584 & 0.657 & 0.618 & 0.736 & 0.567 & 0.640 \\
    \bottomrule
  \end{tabular}
  \caption{Results from evaluating on the 2016 data (not including wikipedia dictionary).}
  \label{table:2016_result}
\end{table*}

\subsection{Neural Network Architecture}
\begin{figure*}
  \centering
  \includegraphics[width=\textwidth]{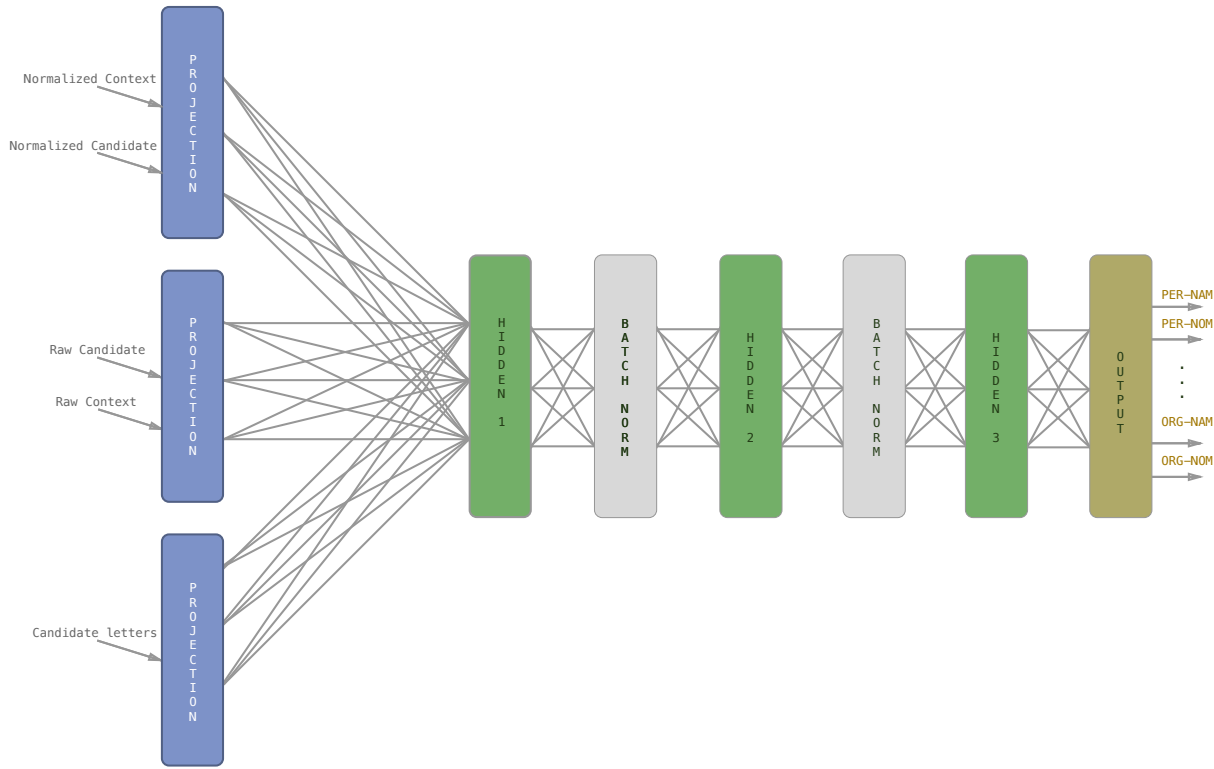}
  \caption{Overview of the full NER network architecture}
  \label{figure:network}
\end{figure*}

The network architecture can be conceptually divided into two parts: A first part projects the input features into a dense space and a second one classifies the input and outputs a class (see Figure~\ref{figure:network}). The classification part of the network consists of three hidden layers, which have batch-normalization layers sliced in-between them, and a final layer that outputs multiclass predictions.

During the development, we tested and evaluated several hyperparameters using a grid search method. Table~\ref{table:hyper_params} shows the final hyperparameters used in the TAC2017 EDL task. We started from initial values identical to those in \citet{Xu2017}.

\begin{table}[h]
  \centering
  \begin{tabular}{l l }
    \hline
    \textbf{Name} & \textbf{Value} \\
    \hline
    Weight initialization & RELU Uniform \\
    Max. window size & 7 \\
    Epoch count & 160\\
    Learning rate & 0.1024 \\
    Dropout & 0.4096 \\
    Optimizer & ADAM \\
    L2 regularization & 0.0 \\
    Neuron count & 512 \\
    Batch size & 512 \\
    Activation function & Leaky RELU \\
    \hline
  \end{tabular}
  \caption{The final hyperparameters used in the TAC2017 EDL task}
  \label{table:hyper_params}
\end{table}

Both the learning rate and dropout followed a linear decay schedule in which they would have a final value of 0.0064 and 0.1024, respectively, by the end of training. We also conducted tests that showed that having a constant, lower dropout rate, yielded slightly better results. We did not use them for the EDL task due to time constraints.

\subsection{Candidate Pruning}
We exhaustively generated all the possible mention candidates that we passed to the classification step. The output is a probability distribution for each class (named and nominal), whose sum is 1. We used the highest probability class to tag the mention if it was 0.5 or greater, otherwise we ignored the output and assigned it to the \texttt{NONE} class.

No overlapping mentions were output, instead each mention had to have no overlap at all. We evaluated two different algorithms to determine which mentions to keep: The highest probability and the longest match:
\begin{itemize}
    \item The highest probability algorithm proceeds from left to right and uses the highest probability, nonoverlapping, leftmost mention.
    \item The longest match instead uses the longest, nonoverlapping, rightmost mention.
\end{itemize} 

During testing, the highest probability algorithm produced the best results, a few points greater than the longest first. The output was also visually cleaner upon manual inspection. We did a grid search for the cutoff value and found that 0.5 produced the best results. Nonetheless, both 0.4 and 0.6 yielded similar results and would be reasonable choices as well.




\section{Entity Linking}

\subsection{Generation of Entity Candidates}
We used the mentions from the recognition step to produce the entity candidates. Each mention found by the FST dictionary has a unique ID that serves as entry point to the mention-entity graph (Sect.~\ref{sec:mention-entity}).


\subsection{Construction of a Local Graph to Disambiguate Entities}

\begin{figure}
  \centering
  \includegraphics[width=0.9\columnwidth]{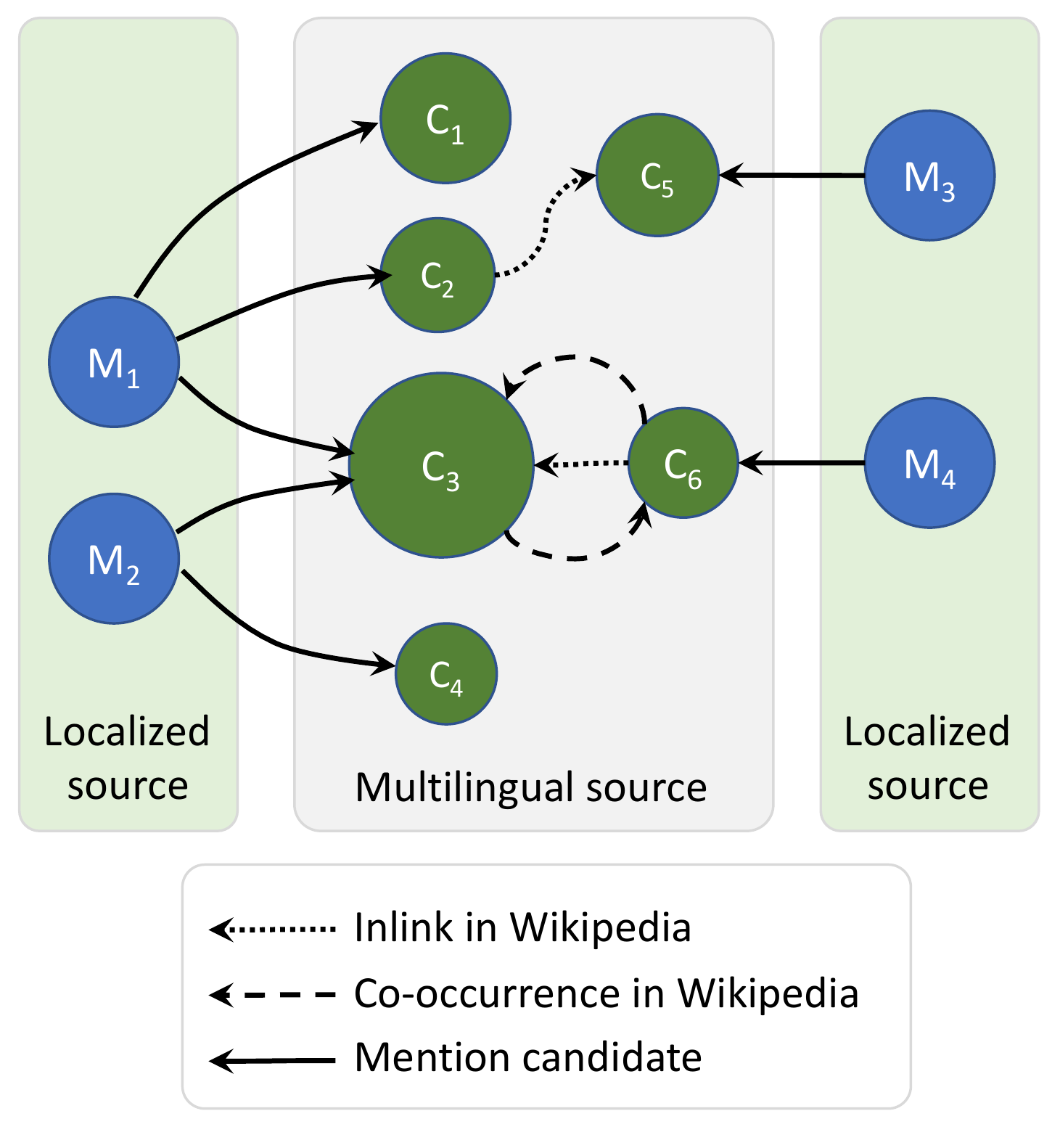}
  \caption{Overview of the entity disambiguation graph}
  \label{figure:disambiguation-graph}
\end{figure}

We build a local graph to disambiguate the entities in a text. The nodes of the graph consist of the mentions and the candidates.  We link these nodes with three types of edges:
\begin{enumerate}
\item The mention-to-candidate edges;
\item The entity cooccurrence edges linking entities when they cooccur in the Wikipedia corpus;
\item The entity inlink edges, reflecting links between pages (entities) in Wikipedia;
\end{enumerate}
While the mentions are language-dependent, the entities reside in a multilingual domain and their edges are aggregated from all the editions of Wikipedia we consider.
Fig.~\ref{figure:disambiguation-graph} shows an overview of the graph.

Following \citet{sodergren2017}, the graph is weighted using the PageRank algorithm \citep{Brin1998}. The candidates are ranked per mention and a normalized weight is produced. The top three candidates are reordered if the candidate title exactly matches the mention.

In contrast to \citet{sodergren2017}, we included the inlinks and we carried out the disambiguation inside a window of 20 mentions. We introduced this window method to reduce the computation and the upper-bound execution time. At the start of the disambiguation, the window is set at the beginning of the text and then shifted by 10 mentions. The windows are then partially overlapping and, in case of conflict, we use the rankings from the left one.

\begin{table*}[htb]
\centering
\begin{tabular}{l|rrr}
\hline
\textbf{Language} & \multicolumn{1}{c}{en} & \multicolumn{1}{c}{es} & \multicolumn{1}{c}{zh} \\ \hline
NERC              & $U_2$: 0.833           & $U_2$: 0.804           & $U_1$/$U_3$: 0.760     \\
NEL               & $U_4$: 0.751           & $U_2$: 0.751           & $U_4$: 0.718           \\
NELC              & $U_4$: 0.726           & $U_2$: 0.733           & $U_1$/$U_3$: 0.760     \\ \hline
CEAFm             & $U_4$: 0.783           & $U_3$: 0.728           & $U_4$: 0.736           \\ \hline
\end{tabular}

\caption{Best version for the named class. F1 score. Results on the TAC 2017 data}
\label{table:named-results}
\end{table*}

\begin{table*}[htb]
\centering
\resizebox{\textwidth}{!}{
\begin{tabular}{l|ccccc|ccccc|ccccc}
\textbf{Languages} & \multicolumn{5}{c|}{en}                                 & \multicolumn{5}{c|}{es}                                 & \multicolumn{5}{c}{zh}                                          \\ \hline
                   & $U_1$ & $U_2$          & $U_3$ & $U_4$          & $U_5$ & $U_1$ & $U_2$          & $U_3$          & $U_4$ & $U_5$ & $U_1$          & $U_2$ & $U_3$          & $U_4$          & $U_5$ \\ \hline
NERC               & 0.813 & \textbf{0.833} & 0.825 & 0.813          & 0.797 & 0.802 & \textbf{0.804} & 0.788          & 0.749 & 0.763 & \textbf{0.760} & 0.750 & \textbf{0.760} & 0.750          & 0.741 \\
NEL                & 0.732 & 0.738          & 0.730 & \textbf{0.751} & 0.691 & 0.747 & \textbf{0.751} & 0.746          & 0.701 & 0.687 & 0.716          & 0.691 & 0.716          & \textbf{0.718} & 0.690 \\
NELC               & 0.711 & 0.717          & 0.710 & \textbf{0.726} & 0.668 & 0.784 & \textbf{0.788} & 0.788          & 0.745 & 0.765 & \textbf{0.757} & 0.755 & \textbf{0.757} & 0.746          & 0.741 \\ \hline
CEAFm              & 0.752 & 0.751          & 0.752 & \textbf{0.783} & 0.754 & 0.722 & 0.714          & \textbf{0.728} & 0.685 & 0.700 & 0.726          & 0.702 & 0.726          & \textbf{0.736} & 0.720
\end{tabular}}

\caption{Breakdown of results for runs $U_1$ to $U_5$, F1 scores, bold indicates the best result per language}
\label{table:breakdown-results}
\end{table*}

\begin{table*}[htb]
\centering
\begin{tabular}{l|ccc|ccc|ccc|ccc|ccc|ccc}
      & \multicolumn{6}{c|}{Mention Recognition}                                                & \multicolumn{3}{c|}{NER} & \multicolumn{9}{c}{Reranker}                                                                                            \\ \hline
      & \multicolumn{3}{c|}{NER-only}              & \multicolumn{3}{c|}{Hybrid}                & \multicolumn{3}{c|}{}    & \multicolumn{3}{c|}{Candidate} & \multicolumn{3}{c|}{Context}               & \multicolumn{3}{c}{None}                  \\ \hline
      & en           & es           & zh           & en           & es           & zh           & en     & es     & zh     & en              & es    & zh   & en           & es           & zh           & en           & es           & zh           \\ \hline
$U_1$ & $\checkmark$ & $\checkmark$ &              &              &              & $\checkmark$ & F      & F      & F      &                 &       &      & $\checkmark$ & $\checkmark$ & $\checkmark$ &              &              &              \\
$U_2$ & $\checkmark$ & $\checkmark$ & $\checkmark$ &              &              &              & F      & F      & F      &                 &       &      & $\checkmark$ & $\checkmark$ & $\checkmark$ &              &              &              \\
$U_3$ &              &              &              & $\checkmark$ & $\checkmark$ & $\checkmark$ & F      & F      & F      & $\checkmark$    &       &      &              & $\checkmark$ & $\checkmark$ &              &              &              \\
$U_4$ &              &              &              & $\checkmark$ & $\checkmark$ & $\checkmark$ & S      & S      & F*     & $\checkmark$    &       &      &              & $\checkmark$ & $\checkmark$ &              &              &              \\
$U_5$ &              &              &              & $\checkmark$ & $\checkmark$ & $\checkmark$ & S      & F*     & S      &                 &       &      &              &              &              & $\checkmark$ & $\checkmark$ & $\checkmark$ \\ \hline
\end{tabular}
\caption{System configuration. F stands for FOFE and S for Stanford NER. F* is an older FOFE model}
\label{table:configuration}
\end{table*}

\subsection{Reranker}
To reduce errors made by the disambiguator and introduce a NIL candidate, we trained a reranker on the TAC 2015-2016 data. We generated a training set of examples by applying the graph-based disambiguator to all the available annotated text. We limited the set of candidates to the top three entities for each mention or up to the correct one if necessary. We then marked each candidate in these lists as positive or negative according to the gold standard. We assigned all the detected mentions overlapping gold standard mentions to the NIL entity. We discarded the rest.

We used two sets of features, \emph{candidate} and \emph{context}, resulting in two models:
\begin{enumerate}
\item The candidate set contains the Jaccard similarity coefficient between the entity title and the mention, the PageRank weight, and the commonness defined as $P(E_{q} | M_{i})$, where $E$ is the entity and $M$ is the specific mention. All the features in the candidate set are encoded as a quantitized one-hot encoded array. 
\item The context set includes the \emph{candidate} features and additional FOFE-encoded left and right contexts surrounding the mention using the same word embeddings as the NER derived from Wikipedia.
\end{enumerate}

We trained the reranker as a binary classifier using a feed-forward neural network with binary cross entropy loss and sigmoid activation. The network consists of 3 dense layers of size 64 for the \emph{candidate} model and 128 for the \emph{context} model.

We incorporated the \emph{candidate} and \emph{context} models into the entity disambiguation using the following equation to produce final ranking score:

\begin{equation}
{\text{Final score}} = {RV} \cdot {RRS}^{\alpha}
\end{equation}
where $RV$ is rank value and $RRS$ is the rerank score, which is equivalent to the prediction probability.

We performed a grid search to find the optimal $\alpha$ for the reranker using the gold standard training set and we selected the best of the two models for each language.

\subsection{Postprocessing}
The postprocessing stage consists of the following steps:  a baseline coreference resolution, a nominal discovery, and a filtering. The baseline coreferencing method uses the linked mentions as input and tries to find exact matches of the first and last word of each linked mention in the text. When such a match is found, the two mentions are coreferring.

To discover the nominal mentions, we first collected a seed word set from the nominal mentions in the TAC data. We then built a dictionary, where the keys were the entities and the values, the nominal phrases compatible with each entity. We extracted these phrases from the Wikidata description of the corresponding entity, as well as the labels and aliases of its instances-of and occupation relations. We finally intersected the resulting set with the seed word set.

Finally, we filtered out the mentions, where we could not find an acceptable class using the entity-to-class mapping dictionary or using the NER predicted class, when available.


\section{Results}


Ugglan was primarily targeting the named entity disambiguation. It was not designed for the nominal and NIL entities, and hence its results on these categories are not at the same level in terms of accuracy. Therefore, we will merely analyze the named results, where the result categories correspond to these acronyms:
\begin{description}
    \item[NERC,] Named Entity Recognition and Classification, corresponding to \texttt{strong\_typed\_mention\_match} in the evaluation script.
    \item[NEL,] Named Entity Linking  (\texttt{strong\_link\_match});
    \item[NELC,] Named Entity Linking and Classification (\texttt{strong\_typed\_link\_match});
    \item[CEAFm,] Clustered Mention Identification (\texttt{CEAFm}).
\end{description}

We submitted five runs, $U_{1}$ to $U_5$. Table~\ref{table:named-results} shows an overview of the best combination per language and type of result taken from the official evaluation data and Table \ref{table:breakdown-results} shows the full breakdown. 

The pipeline setup for the particular runs were selected using the TAC 2016 evaluation as a guide. The runs $U_1$ to $U_3$ used available training data from TAC 2014-2016, while $U_4$ and $U_5$ only used  TAC 2014-2015. Table~\ref{table:configuration} shows the different configurations.


From the results in Table~\ref{table:named-results}, the typed classification is best using only the FOFE-based NER. The Stanford NER is better when it comes to clustered mentions.

\section{Discussion}

\subsection{Mention Recognition}
Ugglan's ability to find linkable mentions is determined by the recall level of the FST dictionary. The NER only helps in reducing noise, thus increasing precision at the expense of possibly lowering the overall recall. The \emph{hybrid} mode tries to mitigate the recall loss by including mentions which have no overlap with any NER mention. NIL mentions are only found using a NER or if the mention was linked and could not be resolved to a Freebase entity. The FOFE NER was trained to identify NOMs, but these were never used as they could not reliably be linked to existing linkable mentions.

\subsection{Linking}
The windowing approach limits the computation complexity at the expense of a possible lower precision. We did not evaluate the effects of the window size and the values were picked arbitrarily using human insight only. Arguably, the optimal size depends on the impact of topic drift, as the linker performs best with a coherent and compatible context with as many related mentions as possible. The more diverse the context is in terms of mentions and candidates, the noisier the graph becomes and the relevant context may shrink as it would require a bigger context to get sufficient supporting candidates to produce a good linkage.

\section*{Acknowledgments}
This research was supported by Vetenskapsr\r{a}det, the Swedish research council, under the \textit{Det digitaliserade samhället} program.

\bibliography{biblio/livre,biblio/lrec,biblio/ned,biblio/nugues,biblio/ref,biblio/bibliography}

\begin{thebibliography}{}
\expandafter\ifx\csname natexlab\endcsname\relax\def\natexlab#1{#1}\fi

\bibitem[{Brin and Page(1998)}]{Brin1998}
Sergey Brin and Lawrence Page. 1998.
\newblock The anatomy of a large-scale hypertextual web search engine.
\newblock {\em Computer Networks\/} 30(1--7):107--117.
\newblock Proceedings of WWW7.

\bibitem[{Finkel et~al.(2005)Finkel, Grenager, and Manning}]{Finkel2005}
Jenny~Rose Finkel, Trond Grenager, and Christopher Manning. 2005.
\newblock Incorporating non-local information into information extraction
  systems by {G}ibbs sampling.
\newblock In {\em Proceedings of the 43nd Annual Meeting of the Association for
  Computational Linguistics (ACL 2005)\/}. Ann Arbor, pages 363--370.

\bibitem[{Ji and Nothman(2016)}]{Ji2016b}
Heng Ji and Joel Nothman. 2016.
\newblock Overview of {TAC-KBP2016} trilingual entity discovery and linking and
  its impact on end-to-end cold-start kbp.
\newblock In {\em Proceedings of the Ninth Text Analysis Conference
  (TAC2016)\/}.

\bibitem[{Ji et~al.(2014)Ji, Nothman, and Hachey}]{Ji2014}
Heng Ji, Joel Nothman, and Ben Hachey. 2014.
\newblock Overview of {TAC-KBP2014} trilingual entity discovery and linking.
\newblock In {\em Proceedings of the Seventh Text Analysis Conference
  (TAC2014)\/}.

\bibitem[{Ji et~al.(2015)Ji, Nothman, Hachey, and Florian}]{Ji2015}
Heng Ji, Joel Nothman, Ben Hachey, and Radu Florian. 2015.
\newblock Overview of {TAC-KBP2015} trilingual entity discovery and linking.
\newblock In {\em Proceedings of the Eighth Text Analysis Conference
  (TAC2015)\/}.

\bibitem[{Ji et~al.(2017)Ji, Pan, Zhang, Nothman, Mayfield, McNamee, and
  Costello}]{Ji2017}
Heng Ji, Xiaoman Pan, Boliang Zhang, Joel Nothman, James Mayfield, Paul
  McNamee, and Cash Costello. 2017.
\newblock Overview of {TAC-KBP2017} 13 languages entity discovery and linking.
\newblock In {\em Proceedings of the Tenth Text Analysis Conference
  (TAC2017)\/}.

\bibitem[{Klang and Nugues(2017)}]{klang2017a}
Marcus Klang and Pierre Nugues. 2017.
\newblock Docforia: A multilayer document model.
\newblock In {\em Proceedings of the 21st Nordic Conference of Computational
  Linguistics\/}. Gothenburg, page 226–230.

\bibitem[{Mikolov et~al.(2013)Mikolov, Yih, and Zweig}]{Mikolov2013c}
Tomas Mikolov, Wen-tau Yih, and Geoffrey Zweig. 2013.
\newblock \href{http://www.aclweb.org/anthology/N13-1090}{Linguistic
  regularities in continuous space word representations}.
\newblock In {\em Proceedings of the 2013 Conference of the North American
  Chapter of the Association for Computational Linguistics: Human Language
  Technologies\/}. Association for Computational Linguistics, Atlanta, Georgia,
  pages 746--751.
\newblock
  \href{http://www.aclweb.org/anthology/N13-1090}{http://www.aclweb.org/anthology/N13-1090}.

\bibitem[{S\"{o}dergren and Nugues(2017)}]{sodergren2017}
Anton S\"{o}dergren and Pierre Nugues. 2017.
\newblock A multilingual entity linker using {PageRank} and semantic graphs.
\newblock In {\em Proceedings of the 21st Nordic Conference of Computational
  Linguistics\/}. Gothenburg, page 87–95.

\bibitem[{Xu et~al.(2017)Xu, Jiang, and Watcharawittayakul}]{Xu2017}
Mingbin Xu, Hui Jiang, and Sedtawut Watcharawittayakul. 2017.
\newblock A local detection approach for named entity recognition and mention
  detection.
\newblock In {\em Proceedings of the 55th Annual Meeting of the Association for
  Computational Linguistics (Volume 1: Long Papers)\/}. Association for
  Computational Linguistics, Vancouver, Canada, pages 1237--1247.

\bibitem[{Zhang et~al.(2015)Zhang, Jiang, Xu, Hou, and Dai}]{Zhang2015}
ShiLiang Zhang, Hui Jiang, MingBin Xu, JunFeng Hou, and LiRong Dai. 2015.
\newblock \href{http://www.aclweb.org/anthology/P15-2081}{The fixed-size
  ordinally-forgetting encoding method for neural network language models}.
\newblock In {\em Proceedings of the 53rd Annual Meeting of the Association for
  Computational Linguistics and the 7th International Joint Conference on
  Natural Language Processing (Volume 2: Short Papers)\/}. Association for
  Computational Linguistics, Beijing, China, pages 495--500.
\newblock
  \href{http://www.aclweb.org/anthology/P15-2081}{http://www.aclweb.org/anthology/P15-2081}.

\end{thebibliography}
\bibliographystyle{acl_natbib}

\end{document}